\documentclass[10pt,twocolumn,letterpaper]{article}

\usepackage{cvpr}
\usepackage{times}
\usepackage{epsfig}
\usepackage{graphicx}
\usepackage{amsmath}
\usepackage{amssymb}
\usepackage{amsthm}
\usepackage{mathrsfs}
\usepackage{bm}
\usepackage{url}
\usepackage{subfigure}

\usepackage[boxruled,lined,linesnumbered]{algorithm2e}
\usepackage{verbatim}

 \cvprfinalcopy 
\def\cvprPaperID{374} 

\ifcvprfinal\fi

 \DeclareMathOperator*{\argmin}{arg~min}

\newcommand{\figlabel}{Figure~}

\newenvironment{narrow}[2]{%
  \begin{list}{}{%
  \setlength{\topsep}{0pt}%
  \setlength{\leftmargin}{#1}%
  \setlength{\rightmargin}{#2}%
  \setlength{\listparindent}{\parindent}%
  \setlength{\itemindent}{\parindent}%
  \setlength{\parsep}{\parskip}}%
\item[]}{\end{list}}

\begin{document}

\title{Modeling Dynamic Swarms}

\author{Bernard Ghanem and Narendra Ahuja\\
University of Illinois at Urbana-Champaign\\
Electrical and Computer Engineering Department\\
{\tt\small \{bghanem2,ahuja\}@vision.ai.uiuc.edu}
}

\maketitle

\begin{abstract}
This paper proposes the problem of modeling video sequences of dynamic swarms (DS). We define DS as a large layout of stochastically repetitive spatial configurations of dynamic objects (swarm elements) whose motions exhibit local spatiotemporal interdependency and stationarity, i.e., the motions are similar in any small spatiotemporal neighborhood.  Examples of DS abound in nature, e.g., herds of animals and flocks of birds. To capture the local spatiotemporal properties of the DS, we present a probabilistic model that learns both the spatial layout of swarm elements and their joint dynamics that are modeled as linear transformations. To this end, a spatiotemporal neighborhood is associated with each swarm element, in which local stationarity is enforced both spatially and temporally. We assume that the prior on the swarm dynamics is distributed according to an MRF in both space and time. Embedding this model in a MAP framework, we iterate between learning the spatial layout of the swarm and its dynamics. We learn the swarm transformations using ICM, which iterates between estimating these transformations and updating their distribution in the spatiotemporal neighborhoods. We demonstrate the validity of our method by conducting experiments on real and synthetic video sequences. Real sequences of birds, geese, robot swarms, and pedestrians evaluate the applicability of our model to real world data. 
\end{abstract}

\section{Introduction}
\begin{figure}[htp]
     \centering
     \begin{narrow}{-0mm}{0mm}
     $\begin{array}{cc}
           \includegraphics[width=.22\textwidth]{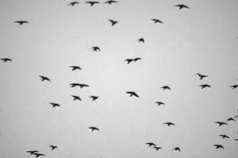}  &   \includegraphics[width=.22\textwidth]{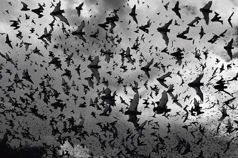} \\
           \includegraphics[width=.22\textwidth]{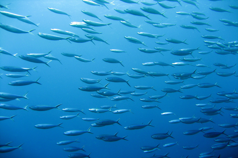} &    \includegraphics[width=.22\textwidth]{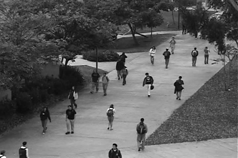}
          \end{array}$
          \end{narrow}
     \caption{Examples of swarms}\label{fig: examples}
\end{figure}

This paper is about modeling of video sequences of a dense collection of moving objects which we will call swarms. Examples of dynamic swarms (DS) in nature abound: a colony of ants, a herd of animals, people in a crowd, a flock of birds, a school of fish, a swarm of honeybees, trees in a storm, and snowfall. In artificial settings, dynamic swarms are illustrated by: fireworks, a caravan of vehicles, sailboats on a lake, and robot swarms.  A DS is characterized by the following properties. (1) All swarm elements belong to the same category. This means that the appearances (i.e. geometric and photometric properties) of the elements are similar although not identical. For example, each element may be a sample from the same underlying probability density function (pdf) of appearance parameters. (2) The swarm elements occur in a dense spatial configuration. Thus, their spatial placement, although not regular, is statistically uniform, e.g., determined by a certain pdf. (3) Element motions are statistically similar. (4) The motions of the swarm elements are globally independent. In other words, the motions of two elements that are sufficiently well separated are independent. However, this is not strictly true on a local scale because if they are located too close compared to the extents of their displacements, then their motions must be interdependent to preserve separation. Thus, the motion parameters of each element vs. the other elements can be considered as being chosen from a mutually conditional pdf. Occasional variations in these swarm properties are also possible, e.g. elements may belong to multiple categories such as different types of vehicles in traffic. Fig. \ref{fig: examples} shows some examples of DS.

This definition of DS is reminiscent of dynamic textures (DT). Indeed, a DS is analogous to a DT of complex nonpoint objects. The introduction of complex nonpoint objects introduces significant complexity: (1) Extraction of nonpoint objects becomes necessary, whose added complexity is evident from, e.g., the algorithm of \cite{AhujaTexelExtractionICCV07}.  (2) Motion for nonpoint objects is richer than point objects, e.g., rotation and nonrigid transformations become feasible. Since most work on DTs has focused on textures formed of pixel or subpixel objects, DS is a relatively unexplored problem. Tools for DS analysis should be useful for general problems such as dynamic scene recognition, dynamic scene synthesis, and anomaly detection, as well as, specific problems such as the motion analysis of animal herds or flocks of birds. In this paper, we present an approach to derive the model of a DS from its video, and demonstrate its efficacy through example applications. Before we do this, we first review the work most related to DS, namely, that on DT.

\subsection*{Related Work}
A DT sequence captures a random spatiotemporal phenomenon which may be the result of a variety of physical processes, e.g., involving objects that are small (smoke particles) or large (snowflakes), or rigid (flag) or nonrigid (cloud, fire), moving in 2D or 3D, etc. Even though the overall global motion of a DT may be perceived by humans as being simple and coherent, the underlying local motion is governed by a complex stochastic model. Irrespective of the nature of the physical phenomena, the objective of DT modeling in computer vision and graphics is to capture the nondeterministic, spatial and temporal variation in images.

As discussed earlier, although the basic notion of DTs allows that both spatial and temporal variations be complex, the limited work done on DT's has focused on moving objects (texels) that have little spatial complexity, even as they exhibit complex motion. The texels are of negligible size (e.g., smoke particles), whose movement appears as a continuous photometric variation in the image, rather than as a sparser arrangement of finite (nonzero) size texels. Consequently, the DT model must mainly capture the motion and less is needed to represent the spatial structure.

Statistical modeling of spatiotemporal interdependence among DT images serves as being closest to the work we present here.
This work includes the spatiotemporal auto-regressive (STAR) model by Szummer et al. \cite{SzummerICIP2004} and multi-resolution analysis (MRA) trees by Bar-Joseph et al. \cite{BarJoseph2001}. The DT model of Doretto et al. \cite{SoattoIJCV2003} uses a stable linear dynamical system (LDS). LDS mixture models have been developed in \cite{ChanPAMI2008} and implemented on DT clustering and segmentation. In \cite{CheBinICME2005}, a mixture of  globally coordinated PPCA models was employed to model a DT.

Along with their merits, the previously proposed models also suffer from certain shortcomings. \textbf{(i)} These models make restrictive assumptions about the DT sequences. Most of them assume that there is either a single DT covering each frame in the sequence. The others that consider multiple DT's are usually limited to particle textures (e.g. water and smoke). Consequently, these models cannot be easily extended to dynamic swarms. Even if the texels were known beforehand, learning a separate model for each texel does not guarantee the underlying spatiotemporal stationarity of DS. \textbf{(ii)} They do not make a clear separation between the appearance and dynamical models of the DT. The approach proposed in \cite{GhanemCVPR08} explicitly aims at this separation, but it is limited to fluid DT's only.

Another body of work that is related to our swarm motion models a DT as a set of dynamic \emph{textons} (or \emph{motons}) whose motion is governed by a Markov chain model \cite{WangZhuPAMI04,WangZhuIJCV05}. This generative model is limited to sequences of particle objects (e.g. snowflakes) or objects imaged at large distances. The texton dynamics are constrained by the underlying assumptions of the model, which state that all textons have the same frame-to-frame transformation, that this transformation is constant over time, and that the dynamics of spatially neighboring textons are independent. While this work does involve moving objects containing more than one pixel per object as well as some interpixel spacing, its modeling power still does not match the needs of the properties (1-4) of a DS given above.

In the rest of this paper, we refer to the objects forming a swarm as swarm \emph{elements}. We propose a probabilistic model that learns both the spatial layout of the swarm elements and their joint dynamics, modeled as linear transformations, which allow for a clear separation between the appearance and dynamics of these elements. This joint representation takes into account the interdependence in the properties of elements that are neighbors in space and time. This is done by enforcing stationarity only within spatiotemporal neighborhoods. This local stationarity constraint allows us to model DS sequences that not only exhibit globally uniform dynamics (to which previous methods are limited), but also sequences whose element properties and dynamics gradually change, in space and time.

\subsection*{Overview of Proposed Model}
Given a DS sequence in which swarm elements undergo locally stationary transformations, we iterate between learning the spatial layout of these elements (i.e. their binary alpha mattes and their frame-to-frame correspondences) and their dynamics. We estimate swarm dynamics such that they follow a probabilistic model that enforces local stationarity within a spatiotemporal neighborhood of each element. In regards to spatial layout, we assume that each swarm element consists of one or more homogenous segments that also possess these spatiotemporal stationarity properties.

We model the frame-to-frame motion of each individual element as a linear transformation, which reconstructs the element's features in a given frame from its features in the previous one. These features can describe local or global properties. In our framework, we do not restrict the choice of these features, since they can be application dependent. These linear transformations are chosen to capture a wide variety of possible changes especially rotation, scaling, and shear. Moreover, a spatiotemporal neighborhood is associated with each element, in which local stationarity is enforced. Spatially, this is done by assuming that the dynamics of elements in a given neighborhood are samples from the same distribution corrupted by i.i.d. Gaussian noise. Temporally, these dynamics are governed by an autoregressive (AR) model. We learn swarm dynamics by estimating the transformations that maximize the a posteriori probability or equivalently that \textbf{(i)} minimize the reconstruction error and \textbf{(ii)} enforce  stationarity in each element's neighborhood.

\vspace{2mm}
\noindent \textbf{Contributions:}
\textbf{(1)} We present an approach that learns the dynamics of swarm elements jointly. This is done by modeling their frame-to-frame linear transformations instead of directly modeling their features. Using these transformations, our model is able to handle more complex swarm motions and allows for a clear separation between the appearance and dynamics of a swarm. \textbf{(2)} Based on our assumption of local spatiotemporal stationarity, the proposed probabilistic model allows for interdependence between swarm elements both in time and space. This is done locally, so as \underline{not} to limit the types of DS sequences that can be modeled, which is a shortcoming of most other methods. \textbf{(3)} The proposed model and learning algorithm estimate the spatial layout of swarm elements by enforcing temporal coherence in determining their frame-to-frame correspondences and the spatial stationarity of their dynamics

\section{Proposed Spatiotemporal Model}\label{sec: proposed model}
In this section, we give a detailed description of our spatiotemporal model for the spatial layout and dynamics of a DS. We consider sequences whose fundamental spatial elements are opaque objects. The changes these elements undergo are stationary, both spatially and temporally. We also assume that each swarm element consists of one or more homogenous segments that also possess these spatiotemporal stationarity properties. To learn the spatial layout of a swarm, we refrain from using texel extraction algorithms (e.g \cite{AhujaTexelExtractionICCV07}) or multiple object trackers from the literature (e.g. \cite{MultipleObjectTracking1}). This is because they do not make use of the spatiotemporal relationship inherent to swarm elements. Instead, we revisit the video segmentation algorithm of \cite{VideoSegmentationICCV09}, which has some interesting properties that we exploit to learn spatial layout. Since no explicit tracking is performed on the swarm elements, occlusion handling remains a problem and is left for future work. To enforce stationarity, we assume that the dynamics of the swarm elements  are distributed according to an MRF in \underline{both} space and time. In our model, the dynamics of each swarm element is influenced by its spatial and temporal neighbors, within its spatiotemporal neighborhood. Unlike other dynamical models (e.g. \cite{SoattoIJCV2003,WangZhuIJCV05}) that assume spatial independence between texture elements, we maintain spatiotemporal dependence among swarm elements to render a more constrained model. In what follows, we give a clear mathematical formulation of our problem.

We are given $F$ frames of size $M\times N$ constituting a swarm sequence. Frame $t$ in this sequence contains $K_t$ swarm elements. This permits that elements can disappear and be formed at different time instances. A swarm element consists of one or more adjacent low-level image segments that have similar dynamics. Note that any low-level segmentation algorithm can be used here. In the following sections, we show how we iterate between learning the spatial layout of the elements and their dynamics. At a given iteration, we fix element dynamics and update the swarm elements by clustering segments to enforce spatiotemporal stationarity. Then, we update the dynamics of the new swarm elements.

Let us denote the swarm elements by their spatial layouts (i.e. binary alpha mattes) $\left\{\mathcal{T}_t^{(i)}\right\}_{t=1,i=1}^{F,K_t}$, where $\mathcal{T}_t^{(i)}$ is the manifestation of the $i^{\text{th}}$ element in frame $t$ and $\mathbb{T}_t=\left\{\mathcal{T}_t^{(i)}\right\}_{i=1}^{K_t}$ is the set of swarm elements in frame $t$.  These swarm elements are represented by their $d$-dimensional feature vectors  $\left\{\vec{f}_t^{(i)}\right\}_{t=1,i=1}^{F,K_t}$, which describe their appearances. To model  local swarm dynamics, we define a linear transformation $\mathcal{A}_t^{(i)}$ that transforms $\vec{f}_t^{(i)}$ into $\vec{f}_{t+1}^{(i)} $. Due to its general form, it can encompass commonly used transformations (e.g. rotation and scaling) as well as more specific ones (e.g. any orthogonal or orthonormal transformation). We use $\mathbb{A}_t=\left\{\mathcal{A}_t^{(i)}\right\}_{i=1}^{K_t}$ to denote the set of transformations for the $K$ elements in frame $t$ and $\mathbb{F}_t=\left\{\vec{f}_t^{(i)}\right\}_{i=1}^{K_t}$ to denote the set of features.

By using frame-to-frame transformations to characterize swarm dynamics instead of their corresponding features, we emphasize the separation between swarm appearance and dynamics. This is usually ignored in other models. This explicit separation allows distinction between and independent control of elements' appearance and motion. That is, we can pair any swarm elements with any dynamics.

The goals of modeling these linear transformations are twofold. \textbf{[G1]} We desire accurate frame-to-frame reconstruction of the feature vectors, which determines how well our model fits the underlying data. \textbf{[G2]} We need to impose spatial and temporal stationarity on the transformations within a local spatiotemporal neighborhood. In the absence of \textbf{[G2]}, our model is ill-posed and too general for any practical use. Consequently, \textbf{[G2]} ensures that our model conforms to the underlying process that generates the swarm elements' dynamics.

Section \ref{subsec: spatiotemporal neighborhood} gives a detailed description of how a swarm element's spatiotemporal neighborhood is formed. In Section \ref{subsec: probabilistics model}, we learn the spatial layout and the linear transformations in a probabilistic MAP framework.

\subsection{Spatiotemporal Neighborhood in a DS}\label{subsec: spatiotemporal neighborhood}
Our dynamical model assumes spatial and temporal stationarity for each swarm element within its spatiotemporal neighborhood.
Let $\mathcal{C}=\left\{\mathcal{N}_t^{(i)}\right\}_{t=1,i=1}^{F,K_t}$ be the set of all spatiotemporal neighborhoods in the sequence. $\mathcal{N}_t^{(i)}$ is the set of elements included in the neighborhood of $\mathcal{T}_t^{(i)}$. We define $\Gamma\left(t,i\right)$  to be the set of index pairs $\left(u,v\right)$ that represent $\mathcal{T}_u^{(v)}$ in $\mathcal{N}_t^{(i)}$. For simplicity, we decompose $\Gamma\left(t,i\right)$ into two disjoint sets of indices, $\Gamma_S\left(t,i\right)$ and $\Gamma_T\left(t,i\right)$, where $\Gamma_S\left(t,i\right)= \left\{(t,j):\mathcal{T}_t^{(j)}\in\mathcal{N}_t^{(i)}\right\}$ and $\Gamma_T\left(t,i\right)= \left\{(s,i):\mathcal{T}_s^{(i)}\in\mathcal{N}_t^{(i)}\right\}$.   $\Gamma_S\left(t,i\right)$ defines the spatial neighbors of $\mathcal{T}_t^{(i)}$, while $\Gamma_T\left(t,i\right)$ defines its temporal neighbors.

\subsubsection*{Spatial Neighborhood} 
The elements, indexed by $\Gamma_S\left(t,i\right)$, are determined by the generalized Voronoi regions corresponding to the elements present in the $t^{\text{th}}$ frame. We also weigh the ``neighborness" of every pair of spatial neighbors. $w_t\left(i,j\right)$ is the corresponding weight for $\left(\mathcal{T}_t^{(i)},\mathcal{T}_{t}^{(j)}\right)$. It is equal to the ratio of the length of the common boundary between the Voronoi regions of the neighboring elements, to the average distance of these elements to the common boundary. For elements that are not spatial neighbors, this weight is set to zero. Local spatial stationarity is enforced by assuming that transformations of neighboring elements are drawn from the same distribution, corrupted by Gaussian i.i.d. noise. Therefore, we have: $\forall~\left(t,j\right)\in\Gamma_S\left(t,i\right):~\mathcal{A}_t^{(j)}=\mathcal{A}_t^{(i)}+N$ where $N(u,v)\sim\mathcal{N}\left(0,\frac{\sigma_S^2}{w_t(i,j)+\varepsilon}\right)~\forall u,v=1,\cdots,d$.

\subsubsection*{Temporal Neighborhood}
The elements, indexed by $\Gamma_T\left(t,i\right)$, are the manifestations of the $i^{\text{th}}$ element in a temporal window consisting of the $W_T$ previous frames. The limits of this window are truncated to remain within the limits of the video sequence itself. This is done to resolve exceptions occurring at the first $W_T$ frames in the sequence.  We enforce temporal stationarity by applying an AR model of order $W_T$ to the sequence of transformations in this temporal window. In fact, the AR model has often been used to model features over time (e.g. \cite{WangZhuPAMI04}), but here, we use it to model the temporal variations of these features (i.e. the dynamics themselves). Therefore, we have  $\forall~\mathcal{N}_t^{(i)}\in\mathcal{C}:~\mathcal{A}_t^{(i)}=\sum_{j=1}^{\rho_t}\alpha_j\mathcal{A}_{t-j}^{(i)}+N$, where $\rho_t=\min\left(W_T,t-1\right)$ and $N(u,v)\sim\mathcal{N}\left(0,\sigma_T^2\right)$. For simplicity, the AR coefficients ($\vec{\alpha}\in\mathbb{R}^{W_T}$), are assumed to be time invariant and the constant for all swarm elements.

\begin{figure*}[htp]\centering
\begin{narrow}{0mm}{0mm}
\centering $\begin{array}{c}
\includegraphics[width=0.98\textwidth]{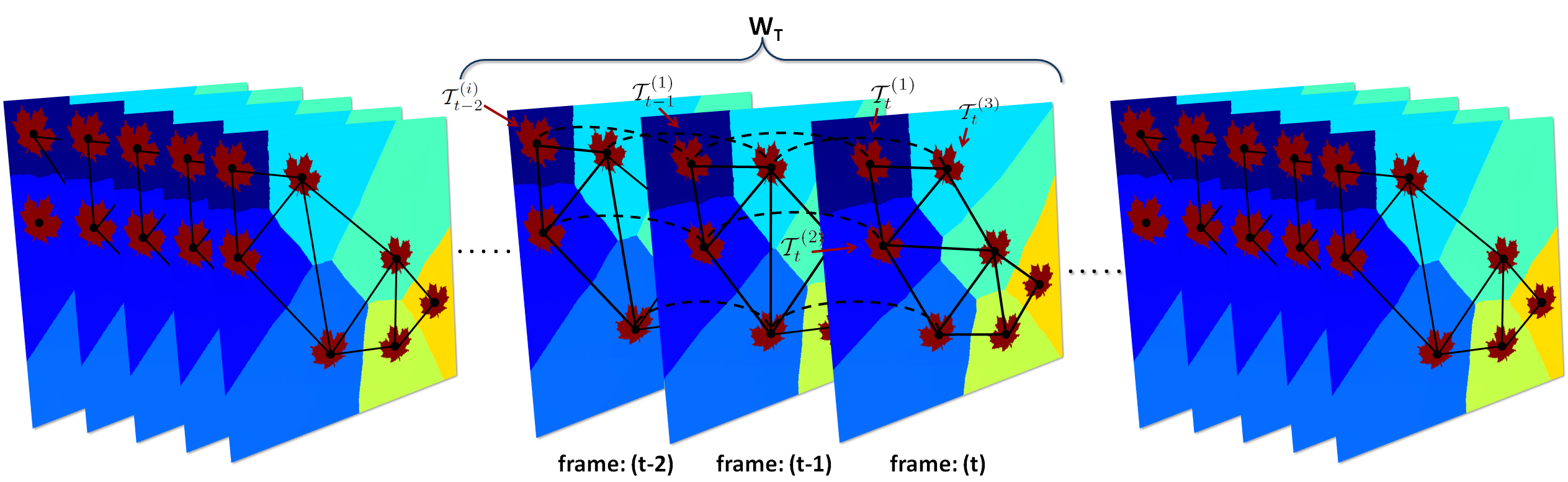}
\end{array}$
\end{narrow}
\caption{Spatial neighbors are connected by solid black lines, while temporal neighbors are connected by dashed black lines. Here, $\mathcal{T}_t^{(1)}$ has two spatial neighbors ($\mathcal{T}_t^{(2)}$ and $\mathcal{T}_t^{(3)}$) and two temporal neighbors
($\mathcal{T}_{t-1}^{(1)}$ and $\mathcal{T}_{t-2}^{(1)}$) comprising its spatiotemporal neighborhood. }\label{fig: spatiotemporal_neighborhood}
\end{figure*}

In Figure \ref{fig: spatiotemporal_neighborhood}, we show an example of the spatiotemporal neighborhood of $\mathcal{T}_t^{(1)}$ with $W_T=2$. Note that the number of spatial neighbors and the ``neighborness" weights can change from frame-to-frame.

\subsection{Model for Swarm Dynamics and Spatial Layout}\label{subsec: probabilistics model}
Here, we present the probabilistic model that governs the dynamics of swarm elements and their spatial layout in a DS. We model the joint probability of the spatial layout of the swarm elements, their features, and their dynamics. This is done by decomposing the joint into the prior over the transformations and the spatial layout, in addition, to the likelihood of the features given the swarm layout and dynamics as in Eq (\ref{eq: joint probability}). In what follows, we model the three terms to ensure \textbf{[G1]} and \textbf{[G2]}.

\begin{align}
&p\left(\left\{\mathbb{A}_t\right\}_{t=1}^{F-1},\left\{\mathbb{F}_t\right\}_{t=1}^{F-1},\left\{\mathbb{T}_t\right\}_{t=1}^{F}\right)=\mathcal{L}\mathcal{P}_{\mathbb{T}}\mathcal{P}_{\mathbb{A}}
\label{eq: joint probability}
\end{align}

\noindent where $\mathcal{L}=p\left(\left\{\mathbb{F}_t\right\}_{t=1}^{F}\mid\left\{\mathbb{A}_t\right\}_{t=1}^{F-1},\left\{\mathbb{T}_t\right\}_{t=1}^{F} \right)$, $\mathcal{P}_{\mathbb{T}}=p\left(\left\{\mathbb{T}_t\right\}_{t=1}^{F} \mid \left\{\mathbb{A}_t\right\}_{t=1}^{F-1}\right)$, and $\mathcal{P}_{\mathbb{A}}=p\left(\left\{\mathbb{A}_t\right\}_{t=1}^{F-1} \right)$.

\subsubsection*{Likelihood Model ($\mathcal{L}$)}
Since we assume a linear relationship between consecutive feature vectors, we can decompose the likelihood probability as: $\mathcal{L}=p_1\prod_{t=1}^{F-1} \prod_{i=1}^{K_t} p\left(\vec{f}_{t+1}^{(i)}\mid\vec{f}_{t}^{(i)},\mathcal{A}_{t}^{(i)},\mathbb{T}_t\right)$, where $\left(\vec{f}_{t+1}^{(i)}\mid\vec{f}_{t}^{(i)},\mathcal{A}_{t}^{(i)},\mathbb{T}_t\right)\sim\mathcal{N}\left(\mathcal{A}_{t}^{(i)}\vec{f}_{t}^{(i)},\gamma_{t}^2I_d\right)$ and $p_1=p\left(\mathbb{F}_1\mid\left\{\mathbb{A}_t\right\}_{t=1}^{F-1},\left\{\mathbb{T}_t\right\}_{t=1}^{F} \right)$ is a constant with respect to the transformations. Consequently, we can write the negative log likelihood as in Eq (\ref{eq: log likelihood}).

\begin{align}
-\text{ln}\left(\mathcal{L}\right)=&\sum_{t=1}^{F-1}\left(\frac{dK_t}{2}\text{ln}\left(\gamma_t^2\right)+\frac{1}{\gamma_t^2}\sum_{i=1}^{K_t}\left\|\vec{f}_{t+1}^{(i)}-
\mathcal{A}_{t}^{(i)}\vec{f}_{t}^{(i)}\right\|_2^2\right)\notag\\
&-\text{ln}\left(p_1\right)+\frac{d\text{ln}\left(2\pi\right)}{2}\sum_{t=1}^{F-1}K_t
\label{eq: log likelihood}
\end{align}

\subsubsection*{Prior on Swarm Spatial Layout ($\mathcal{P}_{\mathbb{T}}$)}
\begin{figure}[htp]\centering
\centering
\includegraphics[width=0.4\textwidth]{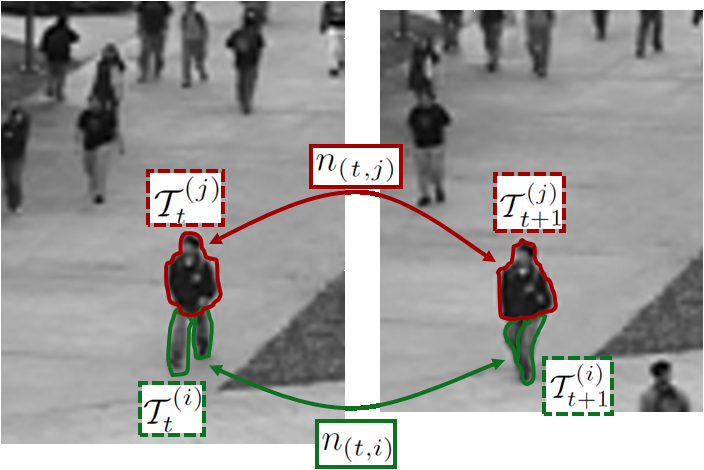}
\caption{Two neighboring nodes of swarm elements in frames $t$ and $t+1$. Note that the $n_{(s,j)}$ consists of two regions.}\label{fig: nodes in swarm graph}
\end{figure}

As stated before, each swarm element consists of one or more homogenous segments that are produced by the algorithm of \cite{EmreSegmentationACCV09}. The spatial layout of these elements and their frame-to-frame correspondences must ensure that the swarm elements' features  are reconstructed faithfully and that spatial stationarity of their dynamics is enforced. The frame-to-frame correspondences of a swarm element are equivalent to many-to-many correspondences between segments from the two frames. To formalize this problem, we denote the frame-to-frame correspondence between $\mathcal{T}_t^{(i)}$ and $\mathcal{T}_{t+1}^{(i)}$ as $n_{(t,i)}$, which is a node in the graph of all frame-to-frame correspondences in the swarm sequence. Two nodes $n_{(t,i)}$ and $n_{(s,j)}$ are considered neighbors in the graph, if any pair of $\left\{\mathcal{T}_t^{(i)},\mathcal{T}_{t+1}^{(i)}\right\}$ and $\left\{\mathcal{T}_s^{(j)},\mathcal{T}_{s+1}^{(j)}\right\}$ are spatially adjacent (i.e. share boundaries). We show an example in Figure \ref{fig: nodes in swarm graph}.

Here, we can define a self-similarity function for each node, $s_1\left(n_{(t,i)}\right)$, that quantifies the quality of frame-to-frame feature reconstruction. Also, we define a pairwise similarity function for each pair of neighboring nodes, $s_2\left(n_{(t,i)},n_{(s,j)}\right)$, that evaluates how similar their frame-to-frame transformations are. This setup is similar to the one used in \cite{VideoSegmentationICCV09}. Actually, we shall see later that we use a similar method to update the spatial layout. We use normalized correlation to define $s_1(.)$ and $s_2(.)$, where $s_1\left(n_{(t,i)}\right)=\frac{\vec{f}_{t+1}^{(i)T}\mathcal{A}_t^{(i)}\vec{f}_t^{(i)}}{\|\vec{f}_{t+1}^{(i)}\|_2\|\mathcal{A}_t^{(i)}\vec{f}_t^{(i)}\|_2}$ and $s_2\left(n_{(t,i)},n_{(s,j)}\right)=\frac{\text{trace}\left(\mathcal{A}_t^{(i)T}\mathcal{A}_s^{(j)}\right)}{\|\mathcal{A}_t^{(i)}\|_F\|\mathcal{A}_s^{(j)}\|_F}$. The prior $\mathcal{P}_{\mathbb{T}}$ is proportional to the self and pairwise similarities of all neighboring nodes in the graph.

\subsubsection*{Prior on Swarm Dynamics ($\mathcal{P}_{\mathbb{A}}$)}
As $\mathcal{L}$ was modeled to guarantee \textbf{[G1]}, \textbf{[G2]} is accounted for by modeling $\mathcal{P}_{\mathbb{A}}$ as a product of potential functions defined on the set of all spatiotemporal neighborhoods. This decomposition is widely used to model priors on maximum cliques defined on an undirected graph. We define the potential function for each clique as the product of a spatial potential $\Psi_S(.)$ and a temporal  potential $\Psi_T(.)$, which guarantee spatial and temporal stationarity in swarm dynamics, respectively. So, we have $\mathcal{P}_{\mathbb{A}}=\frac{1}{Z}\prod_{\mathcal{N}_t^{(i)}\in\mathcal{C}}\left[\Psi_S\left(\mathcal{N}_t^{(i)}\right)\Psi_T\left(\mathcal{N}_t^{(i)}\right)\right]$, where

\noindent$\begin{cases}
\Psi_S\left(\mathcal{N}_t^{(i)}\right)&=\prod_{(t,j),(t,j^{'})\in\Gamma_S(t,i):j\neq j^{'}}f_S\left(\mathcal{A}_t^{(j)},\mathcal{A}_t^{(j^{'})}\right)\\
\Psi_T\left(\mathcal{N}_t^{(i)}\right)&=f_T\left(\left\{\mathcal{A}_s^{(i)}: \mathcal{T}_s^{(i)}\in\mathcal{N}_t^{(i)}\right\}\right)
\end{cases}$

\noindent $f_S$ and $f_T$ are potentials that evaluate how spatially and temporally stationary the swarm transformations are.  For simplicity, we set $f_S\left(\mathcal{A}_t^{(j)},\mathcal{A}_t^{(j^{'})}\right)=p\left(\mathcal{A}_t^{(j)}\mid\mathcal{A}_t^{(j^{'})}\right)$ and
$\Psi_T\left(\mathcal{N}_t^{(i)}\right)=p\left(\mathcal{A}_t^{(i)}\mid\left\{\mathcal{A}_{t-j}^{(i)}\right\}_{j=1}^{\rho_t}\right)$. We can express the negative log prior as in Eq (\ref{eq: log prior}). Note that $p_2$ is a constant that depends on the ``neighborness" weights, $C_S=\sum_{\mathcal{N}_t^{(i)}\in\mathcal{C}}\frac{d^2}{2}\left|\Gamma_S(t,i)\right|^2$ , and $C_T=\sum_{\mathcal{N}_t^{(i)}\in\mathcal{C}}\frac{d^2}{2}\left|\Gamma_T(t,i)\right|^2$. Also, we assume that the normalizing factor $Z$ is  constant with respect to the swarm dynamics, the noise variances, and the AR coefficients.

\begin{align}
&-\text{ln}\left(\mathcal{P}_{\mathbb{A}}\right) =\text{ln}\left(Z\right)+\text{ln}\left(p_2\right)+C_S\left[\text{ln}\left(\sigma_S^2\right)\right]+ C_T\left[\text{ln}\left(\sigma_T^2\right)\right]\notag\\
&+\frac{1}{\sigma_S^2}\sum_{\mathcal{N}_t^{(i)}\in\mathcal{C}}
\left(\sum_{(t,j),(t,j^{'})\in\Gamma_S(t,i)}w_t(j,j^{'})\left\|\mathcal{A}_t^{(j)}-\mathcal{A}_t^{(j^{'})}\right\|_F^2\right)\notag\\
&+\frac{1}{\sigma_T^2}\sum_{\mathcal{N}_t^{(i)}\in\mathcal{C}}\left\|\mathcal{A}_t^{(i)}-\sum_{j=1}^{\rho_t}\alpha_j\mathcal{A}_{t-j}^{(i)}\right\|_F^2\label{eq: log prior}
\end{align}

\subsection{Learning Swarm Layout and Dynamics}\label{subsec: probabilistics model}
After establishing our probabilistic model, we proceed to learning its parameters,  $\left\{\mathbb{T}_t\right\}_{t=1}^{F}$, $\left\{\mathbb{A}_t\right\}_{t=1}^{F-1}$, the noise variances $\sigma_S$, $\sigma_T$, and $\vec{\gamma}$ (i.e. $\left\{\gamma_t\right\}_{t=1}^{F-1}$), as well as the AR coefficients $\vec{\alpha}$ (i.e. $\left\{\alpha_j\right\}_{j=1}^{W_T}$).  To do this, we embed our model into a MAP framework. We assume that the prior on the features and the prior on the noise variances are uniform. Replacing Eq (\ref{eq: log likelihood},\ref{eq: log prior}) in Eq (\ref{eq: joint probability}), we formulate the MAP problem as a nonlinear and non-convex minimization problem.

\begin{align}
\underset{\left\{\mathbb{T}_t\right\}_{t=1}^{F},\left\{\mathbb{A}_t\right\}_{t=1}^{F-1},~\sigma_S,~\sigma_T,~\vec{\gamma},\vec{\alpha}}{\min}~\left[\left(-\text{ln}\mathcal{L}-\text{ln}\mathcal{P}_{\mathbb{A}}\right)-\text{ln}\mathcal{P}_{\mathbb{T}}\right]\label{eq: MAP rule}
\end{align}

\noindent Due to the complex form of Eq (\ref{eq: MAP rule}), we learn the spatial layout of the DS and its dynamics in an iterative fashion. In each iteration, we either fix the dynamics and update the spatial layout or vice versa. In what follows, we show the steps involved in updating the spatial layout and the dynamics at the $j^{\text{th}}$ iteration.

\subsubsection*{Spatial Layout Update}
We employ a method similar to the one used for video object segmentation in \cite{VideoSegmentationICCV09} to update $\left\{\mathbb{T}_t[j-1]\right\}_{t=1}^{F}$. We will only highlight the main aspects of this method and how it applies to modeling DS's. We create a graph whose nodes are all candidates for frame-to-frame correspondences between $\left\{\mathbb{T}_t[j-1]\right\}_{t=1}^{F}$ and individual segments of these frames. In other words, a segment or swarm element in frame $t$ corresponds to a segment or swarm element in the next frame, if the projection of the former into frame $t+1$ (according to its optical flow) overlaps with the latter. This graph allows for the clustering of similar and neighboring nodes, thus, enabling many-to-many correspondences between consecutive frames. Once this graph is created, the attributes of each node and the edge weights between neighboring nodes are determined by $s_1(.)$ and $s_2(.)$, as defined in Section \ref{subsec: probabilistics model}. For segments that do not belong to $\left\{\mathbb{T}_t[j-1]\right\}_{t=1}^{F}$, we use identity for their transformation. Given this weighted undirected graph, we cluster its nodes into valid and invalid correspondences. This binary clustering is done using graph cuts, instead of relaxation labeling. Then, the resulting valid correspondences are broken down into individual connected components, where connectedness is over time and space. This yields $\left\{\mathbb{T}_t[j]\right\}_{t=1}^{F}$. As pointed out in \cite{VideoSegmentationICCV09}, this method tends to cluster adjacent/occluding swarm elements with similar dynamics. For initialization, we set $\left\{\mathbb{T}_t[0]\right\}_{t=1}^{F}$ to all segments in the video sequence with non-zero optical flow.

\subsubsection*{Dynamics Update}
Given  $\left\{\mathbb{T}_t[j]\right\}_{t=1}^{F}$, Eq. (\ref{eq: MAP rule}) can be solved iteratively using Iterated Conditional Modes (ICM) \cite{BesagICM1986}, which guarantees a local minimum. In the $k^{\text{th}}$ ICM iteration, the variances are updated to their ML estimates. Updating each $\mathcal{A}_t^{(i)}$ requires the minimization of a convex quadratic, matrix problem. $\vec{\alpha}$ is updated by solving a linear system of equations. In what follows, we index the model parameters with $[k]$ to denote their estimates in the $k^{\text{th}}$ ICM iteration.

First, we show the update equation for the AR coefficients. Taking the gradient of Eq (\ref{eq: MAP rule}) with respect to $\vec{\alpha}$ and setting it to zero renders the following update equation: $\mathcal{M}\vec{\alpha}[k]=\vec{m}$.  Here, $\mathcal{M}$ is the sum of Gramm matrices corresponding to the transformations associated with the spatiotemporal neighborhoods at iteration $k$. $\vec{m}$ is the sum of the inner products between these transformations.


Now, we turn to updating the transformations. At each ICM iteration, we fix all of them except for $X=\mathcal{A}_t^{(i)}[k]$. Here, we isolate the dependence of Eq (\ref{eq: MAP rule}) on $X$ and minimize the following convex-quadratic matrix problem.

\begin{align}
\min_{}~g\left(X\right)=\frac{e_R\left(X\right)}{\gamma_t^2[k]}+\frac{2e_S\left(X\right)}{\sigma_S^2[k]}+\frac{e_T\left(X\right)}{\sigma_T^2[k]}
\label{eq: transformation update problem}
\end{align}

\noindent where $e_R$ and $e_S$ represent the reconstruction and spatial stationarity residuals, respectively. $e_T$ represents the temporal stationarity residuals corresponding to the frames preceding frame $t$. We express these terms as follows.


\begin{align}
\begin{cases}
e_R\left(X\right)=\left\|\vec{f}_{t+1}^{(i)}-X\vec{f}_t^{(i)}\right\|_2^2\\
e_S\left(X\right)=\sum_{(t,i^{'})\in\Gamma_S(t,i)}w_t(i,i^{'})\left\|X-\mathcal{A}_t^{(i^{'})}[k]\right\|_F^2\\
e_T\left(X\right)=\left\|X-\sum_{j=1}^{\rho_t}\alpha_j\mathcal{A}_{t-j}^{(i)}\right\|_F^2+ \notag\\
\sum_{k=1}^{\min(W_T,F-t)}\left\|\alpha_k X-\mathcal{A}_{t+k}^{(i)}[k]+\sum_{\underset{j\neq k}{j=1}}^{\rho_{t+k}}\alpha_j\mathcal{A}_{t+k-j}^{(i)}[k]\right\|_F^2 \notag
\end{cases}
\end{align}

Minimizing $g\left(X\right)$ is a convex quadratic problem that admits a global minimum $X^{*}$. It can be obtained using gradient descent where the rate of descent ($\eta$) is determined by a line search. A closed form solution for $\eta$ can be derived. Till now, $X$ has been an unconstrained linear transformation; however, certain applications require that it belong to a feasible set $\mathbb{S}_d$ (e.g. rotation or symmetric matrices). To do this, we project the intermediate solution at each descent step onto $\mathbb{S}_d$. In some cases, this projection is trivial. For example, if $\mathbb{S}_d=\left\{X\in\mathbb{R}^{d\times d}:X=X^T\right\}$, the projection of $X$ is $\frac{X+X^T}{2}$. Using differential matrix identities, we can express the gradient of $g\left(X\right)$ in a computationally efficient form: $\nabla g=X\left(\beta I_d+\vec{b}\vec{b}^T\right)-D$ where $\beta$, $\vec{b}$,  and $D$ are functions of $\vec{f}_t^{(i)}$, $\vec{f}_{t+1}^{(i)}$, and the current estimates of the transformations and $\vec{\alpha}$. \textbf{Algorithm \ref{algo: gradient descent}} provides details for solving Eq (\ref{eq: transformation update problem}). 


We can initialize $X$ in two ways. \textbf{(a)} Set $X_{(0)}$ equal to the transformation obtained from the previous ICM iteration (i.e. $X_{(0)}=\mathcal{A}_t^{(i)}[k-1]$). \textbf{(b)} If $X$ is constrained to be in $\mathbb{S}_d$, we can initialize $X_{(0)}$ by projecting the solution  to the unconstrained version of Eq (\ref{eq: transformation update problem}), denoted $X_{\text{UNC}}^*$, onto $\mathbb{S}_d$.  Setting $\nabla g=0$ and using the  matrix inversion lemma, we get $X_{\text{UNC}}^*=\frac{D}{\beta}\left[I_d-\frac{\vec{b}\vec{b}^T}{\beta+\|\vec{b}\|_2^2}\right]$. In our experiments, both initialization schemes had similar rates of convergence; however, \textbf{(b)} tends to be more numerically unstable when $\beta$ is small. For the first ICM iteration ($k=0$), we initialize every $\mathcal{A}_t^{(i)}[0]=\text{\textbf{0}}_d$. Numerically, we avoid division by zero by setting $\sigma_S[0]=\sigma_T[0]=\gamma_t[0]=1$.

\begin{algorithm}[h!]
\label{algo: gradient descent} \caption{\emph{Gradient Descent} (\textbf{GD}) }
\SetKwInOut{Input}{Input} \SetKwInOut{Output}{Output}

\Input{$X_{(0)}\in\mathbb{S}_d$, $\beta$, $\vec{b}$, $D$, $\epsilon$}
\BlankLine

\textbf{Initialization}: $\delta\leftarrow\infty$; $\ell=0$

\While{$\delta\geq\epsilon$}{
$\eta_{\ell}=\argmin_{\eta\geq0}g\left(X_{(\ell)}-\eta \left(\nabla g\right)|_{X_{(\ell)}}\right)$\\\BlankLine

$X_{\left(\ell+\frac{1}{2}\right)}=X_{(\ell)}-\eta_l \left(\nabla g\right)|_{X_{(\ell)}}$\\

$X_{\left(\ell+1\right)}=\text{\textbf{P}}_{\mathbb{S}_d}\left[X_{\left(\ell+\frac{1}{2}\right)}\right]$ ~~~\textbf{(optional)}
\BlankLine
$\delta = \frac{\|X_{(\ell+1)}-X_{(\ell)}\|_F}{\|X_{(\ell)}\|_F}$;
$\ell=\ell+1$
}

\end{algorithm}

\textbf{Algorithm \ref{algo: texel dynamics}} combines all these update equations together into the overall algorithm for solving Eq (\ref{eq: MAP rule}) to learn the swarm spatial layout and dynamics. The worst case complexity of this algorithm is $\mathcal{O}(Fd^3)$, since it is defined by the complexity of \textbf{Algorithm \ref{algo: gradient descent}}  that has a linear convergence rate.

\begin{algorithm}[h!]
\label{algo: texel dynamics} \caption{\emph{Learn Swarm Layout and Dynamics}}
\SetKwInOut{Input}{Input} \SetKwInOut{Output}{Output}

\Input{$\left\{\mathbb{F}_t, \mathbb{T}_t[0],\mathbb{A}_t[0]\right\}_{t=1}^{F}$, $W_T$, $\epsilon$, $j_{\text{max}}$, $k_{\text{max}}$}
\BlankLine
\For{$j \leftarrow 0$ \textbf{TO} $j_{\text{max}}$}{
\emph{// update spatial layout}\\
$\bullet$ get $\left\{\mathbb{T}_t[j+1]\right\}_{t=1}^{F}$ from $\left\{\mathbb{T}_t[j],\mathbb{A}_t[j]\right\}_{t=1}^{F}$

\For{$t\leftarrow 1$ \textbf{TO} $F$; $i\leftarrow 1$ \textbf{TO} $K_t$}{
$\bullet$  find generalized Voronoi regions of $\mathcal{T}_t^{(i)}$\\
$\bullet$  compute $w_t(t,i)$
}

\emph{// update noise variances and transformations}\\
\textbf{Initialization}: $\delta\leftarrow\infty$; $k=0$\\
\While{$\left(\delta\geq\epsilon\right)$ \textbf{AND} $\left(k\leq k_{\text{max}}\right)$ }{

$\bullet$~compute $\sigma_S[k]$, $\sigma_T[k]$, $\vec{\gamma}[k]$, $\vec{\alpha}[k]$\\

\BlankLine
\For{$t\leftarrow 1$ \textbf{TO} $F$; $i\leftarrow 1$ \textbf{TO} $K_t$}{
$\bullet$~compute $\beta$, $\vec{b}$, $D$, $X_{(0)}$\\
$\bullet~\mathcal{B}_t^{(i)}[k+1]$=\textbf{GD}$\left(X_{(0)},\beta,\vec{b},D,\epsilon\right)$
}

$\delta =\max_{(t,i)}\frac{\|\mathcal{A}_t^{(i)}[k+1]-\mathcal{A}_t^{(i)}[k]\|_F}{\|\mathcal{A}_t^{(i)}[k]\|_F};$
$k= k+1$\\
$\bullet~\mathcal{A}_{t}^{(i)}[j]=\mathcal{B}_t^{(i)}[k+1] ~\forall t,i$
}
}
\end{algorithm}

\section{Experimental Results}\label{sec: experimental results}
To validate our model and evaluate the performance of our algorithm, we conducted experiments on synthetic sequences (Section (\ref{subsec: exp_synthetic})) and real sequences (Section (\ref{subsec: exp_real})). The synthetic sequences help provide quantitative evaluation. The experiments show that we can learn the dynamics of swarms and discriminate between different types of swarm motion.

\subsection{Synthetic Sequences}\label{subsec: exp_synthetic}
\begin{figure}[htp]
     \centering
     \subfigure[sample frame]{\label{subfig: sample frame}
          \includegraphics[width=.23\textwidth]{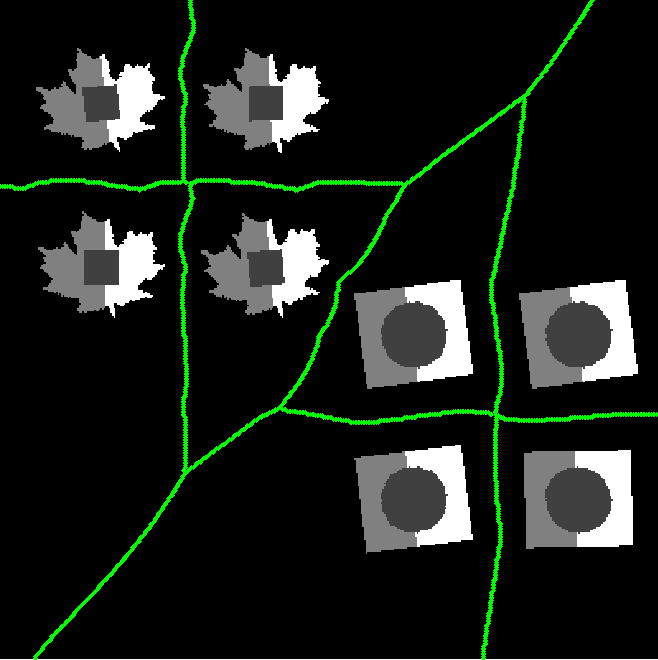}}
     \subfigure[learned transformation: $\mathcal{A}_{10}^{(1)}$]{\label{subfig: sample transformation}
          \includegraphics[width=.21\textwidth]{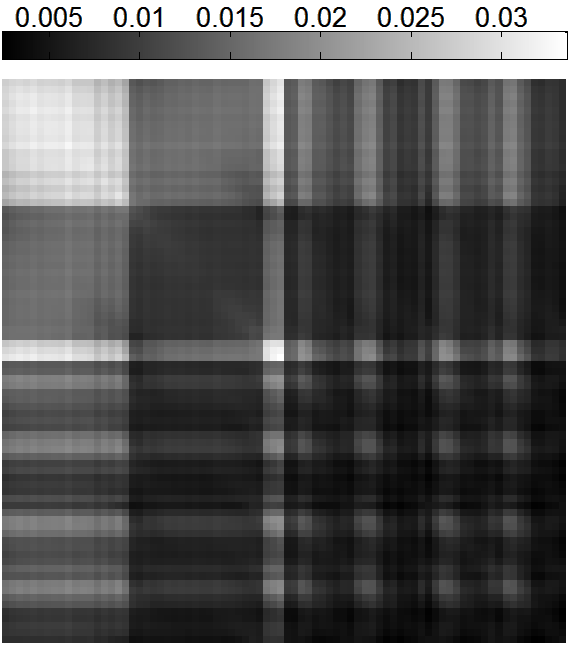}} \\
     \subfigure[modeling performance]{\label{subfig: performance graphs}
         \fbox{ \includegraphics[width=.42\textwidth]{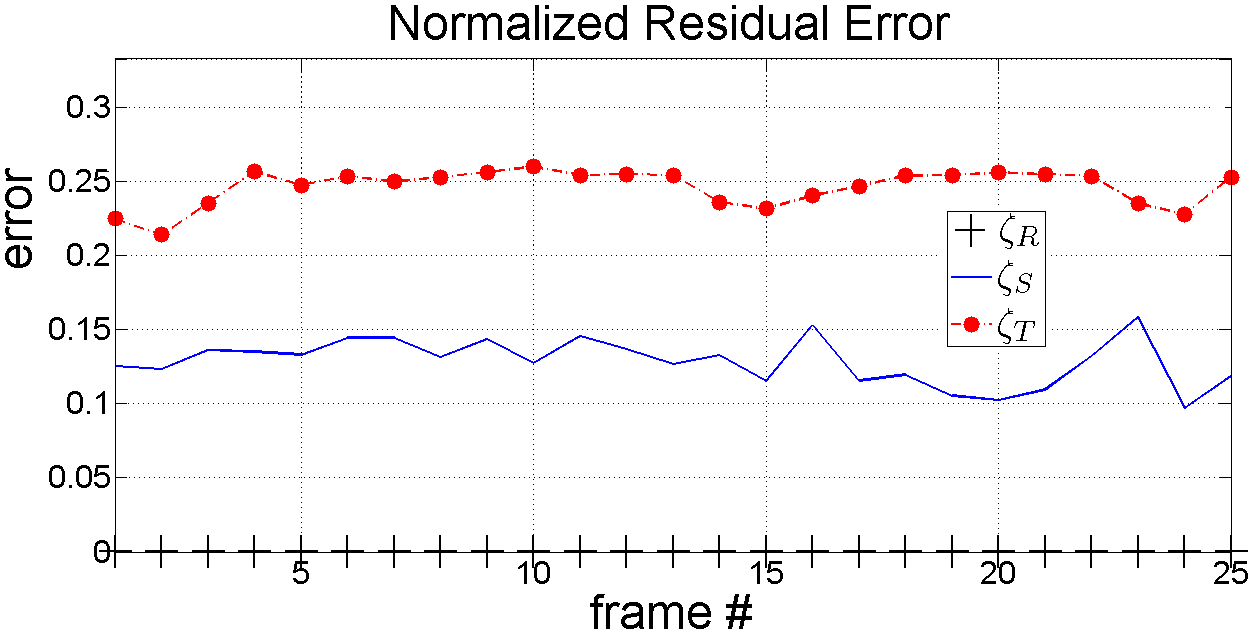}}}

     \caption{\ref{subfig: sample frame} is a frame in the synthetic sequence. \ref{subfig: sample transformation} shows transformation $\mathcal{A}_{10}^{(1)}$, after convergence. All the video results are provided in the \textbf{supplementary material}.}
\end{figure}

\noindent \textbf{Model Learning:} First, we construct a synthetic DS sequence of $F=25$ frames and $K=8$ elements ($4$ leaves and $4$ squares with a simple textured interior). Figure \ref{subfig: sample frame} shows a sample frame of this sequence, where the boundaries of the generalized Voronoi regions are drawn in green. The motion of the swarm elements is synthesized by applying a globally similar rotation $\mathcal{R}_{\theta_t^{(i)}}$. Specifically, for each element in every frame, $\theta_t^{(i)}$ is sampled from a Gaussian distribution $\mathcal{N}(\theta_0=\frac{\pi}{25},\sigma=\frac{1}{50})$

The features we used were based on a polar coordinate system centered at the centroid of each element, where each angular bin had a width of $\frac{\pi}{20}$ rad. For each angular bin, we extracted two shape features (kurtosis and skew), the mean centroidal distance of the element boundary, and the mean intensity value. This yielded a feature vector of size $d=160$. Setting $\epsilon=10^{-3}$, $k_{\text{max}}=50$ and $W_T=3$, we applied \textbf{Algorithm \ref{algo: texel dynamics}} to learn the swarm dynamics. Running MATLAB on a 2.4GHz PC, our algorithm converged in $40$ ICM iterations ($\sim 30$ seconds). Figure \ref{subfig: sample transformation} shows a sample transformation matrix after convergence. We evaluate our model fitting performance by using three measures: the reconstruction residual error $\zeta_R(t)$, the spatial residual error $\zeta_S(t)$, and the temporal residual error $\zeta_T(t)$ defined as:

\begin{align}
\begin{cases}
\zeta_R(t)=\frac{1}{K}\sum_{i=1}^K \frac{1}{\|\vec{f}_t^{(i)}\|_2} \sqrt{e_R\left(\mathcal{A}_t^{(i)}\right)}\\
\zeta_S(t)=\frac{1}{K}\sum_{i=1}^K \frac{1}{|\Gamma_S(t,i)|\|\mathcal{A}_t^{(i)}\|_F} \sqrt{e_S\left(\mathcal{A}_t^{(i)}\right)}\\
\zeta_T(t)=\frac{1}{K}\sum_{i=1}^K \frac{1}{|\Gamma_T(t,i)|\|\mathcal{A}_t^{(i)}\|_F} \sqrt{e_T\left(\mathcal{A}_t^{(i)}\right)}
\end{cases} \notag
\end{align}

They quantify the average error incurred in reconstructing the data and enforcing stationarity in the spatiotemporal neighborhood of each swarm element. Clearly, the smaller these measures are, the better our model fits the data. Figure \ref{subfig: performance graphs} plots these measures for all frames in the sequence. All three measures show a stable variation with time. $\zeta_S$ and $\zeta_T$ are consistently larger than $\zeta_R$ due to the added noise corrupting each transformation. In fact, as $\sigma\rightarrow  0$, $\zeta_S$ and $\zeta_T$ both get closer to $\zeta_R$. Furthermore, $\zeta_T$ is consistently larger than $\zeta_S$ because temporal neighborhoods only extend $W_T=3$ frames from each swarm element. In fact, as $W_T\rightarrow \left( F-1\right)$, $\zeta_T$ gets closer to $\zeta_S$, since temporal stationarity is enforced on a larger number of frames. Here, we point out that although the leaf and square elements are significantly different in appearance, their dynamics are the same. This reinforces the fact that our method successfully separates between swarm appearance and dynamics.

\begin{figure}[htp]
     \centering
     \begin{narrow}{-6mm}{0mm}
     $\begin{array}{cc}
          \includegraphics[width=.23\textwidth]{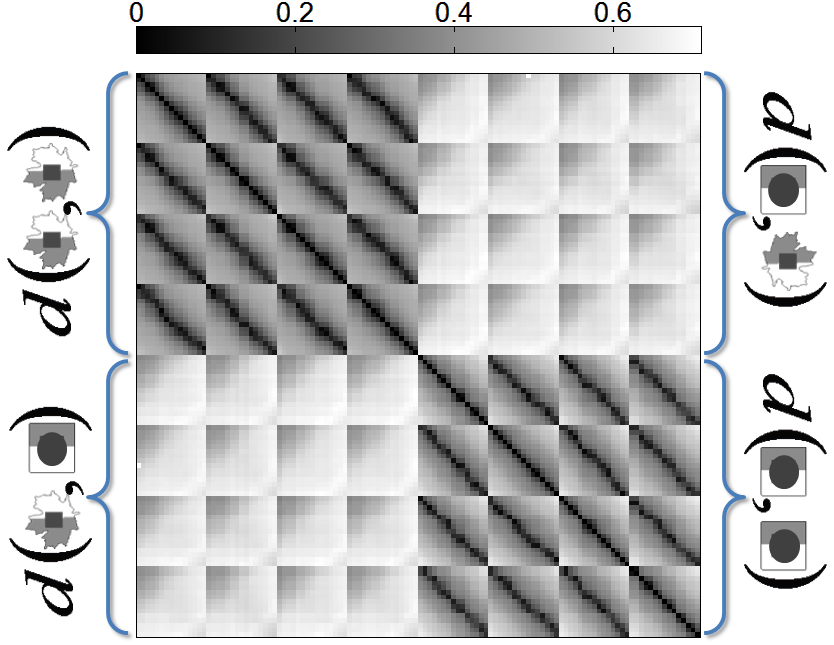}  &   \fbox{ \includegraphics[width=.24\textwidth]{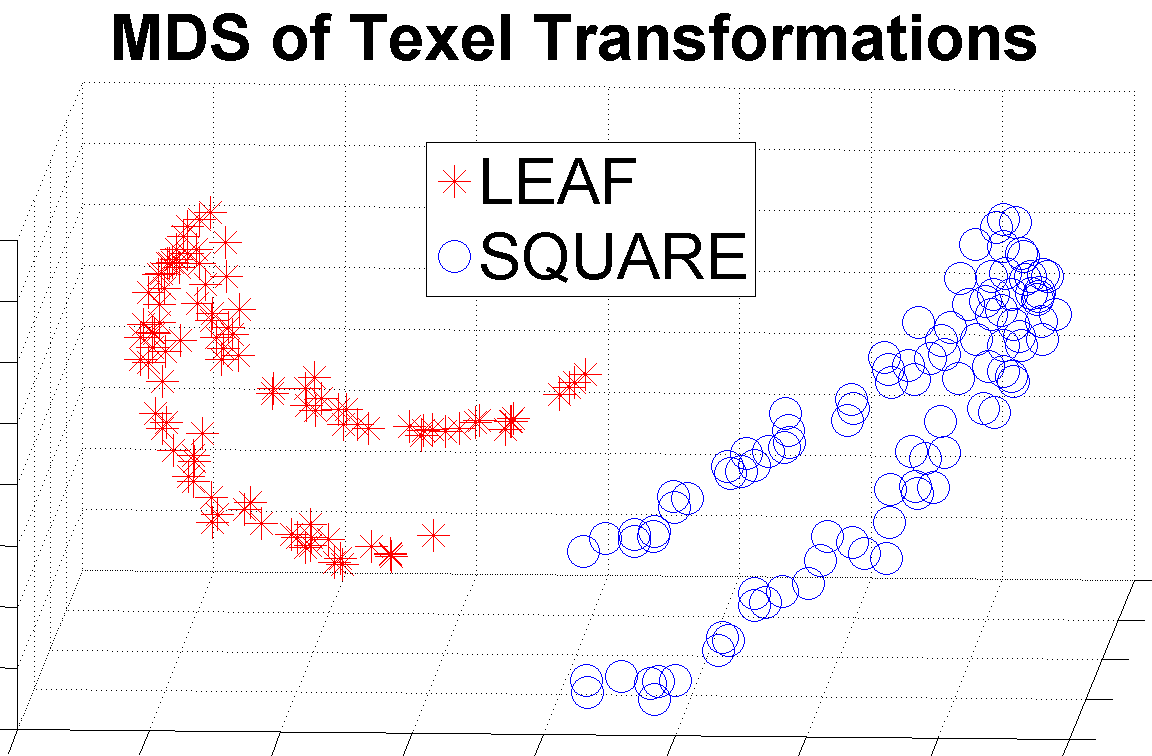} }       \\
          \textrm{(a) distance matrix} & \textrm{(b) MDS of swarm dynamics}
          \end{array}$
          \end{narrow}

          \begin{narrow}{-3mm}{-2mm}
     \caption{\ref{fig: discrimination}(a) shows the distances between the swarm transformations in the synthetic sequence. Note that brighter values designate larger distances. \ref{fig: discrimination}(b) projects the transformations onto $\mathbb{R}^3$ using MDS. }\label{fig: discrimination}\end{narrow}
\end{figure}

\vspace{2mm}
\noindent \textbf{Motion Discrimination:} Here, we demonstrate that the learned transformations can discriminate between different types of motion. Another synthetic DS sequence is constructed in the same manner as before, but with the leaf and square elements now rotating in \underline{opposite} directions. Leaf elements undergo $\mathcal{R}_{\theta_t^{(i)}}$, while square elements undergo $\mathcal{R}_{-\theta_t^{(i)}}$. After learning the swarm dynamics, we compute all the distances (i.e. Frobenius norm of the difference) between pairs of learned transformations. We show the resulting distance matrix in Figure \ref{fig: discrimination}(a). We see that the transformations corresponding to the leaf elements are close to each other and far from those corresponding to the square elements. For visualization purposes, we perform MDS on these pairwise distances to embed the transformations in $\mathbb{R}^3$. In this space, the leaf and square dynamics are easily separable. Moreover, these transformations can be perfectly clustered using spectral clustering ($K=2$).

This result reinforces the fact that our method can successfully learn and discriminate between different motions occurring within a single DS sequence. This conclusion is valid as long as the ``neighborness" weights associated with  swarm elements undergoing similar dynamics are reasonably higher than those moving differently.

\subsection{Real Sequences}\label{subsec: exp_real}
In this section, we present experimental results produced when \textbf{Algorithm} \ref{algo: texel dynamics} is applied to real sequences where single or multiple elements are undergoing an underlying dynamic swarm motion.

\subsubsection{Single Swarm Element Sequences} 
Here, we apply our algorithm to human action recognition, where we consider the human as a single texel. There is no need to determine the spatial neighborhoods of the texels. The action sequences were obtained from the Weizmann classification database \cite{WeizmannHumanActionDatabase}, which contains $10$ human actions. We use background subtraction to extract the texels. In addition to the features used earlier, we use the height and the width of the texel masks at each frame.

\begin{figure}
     \centering
     \begin{narrow}{-4mm}{-6mm}
     $\begin{array}{cc}
          \fbox{\includegraphics[width=.27\textwidth]{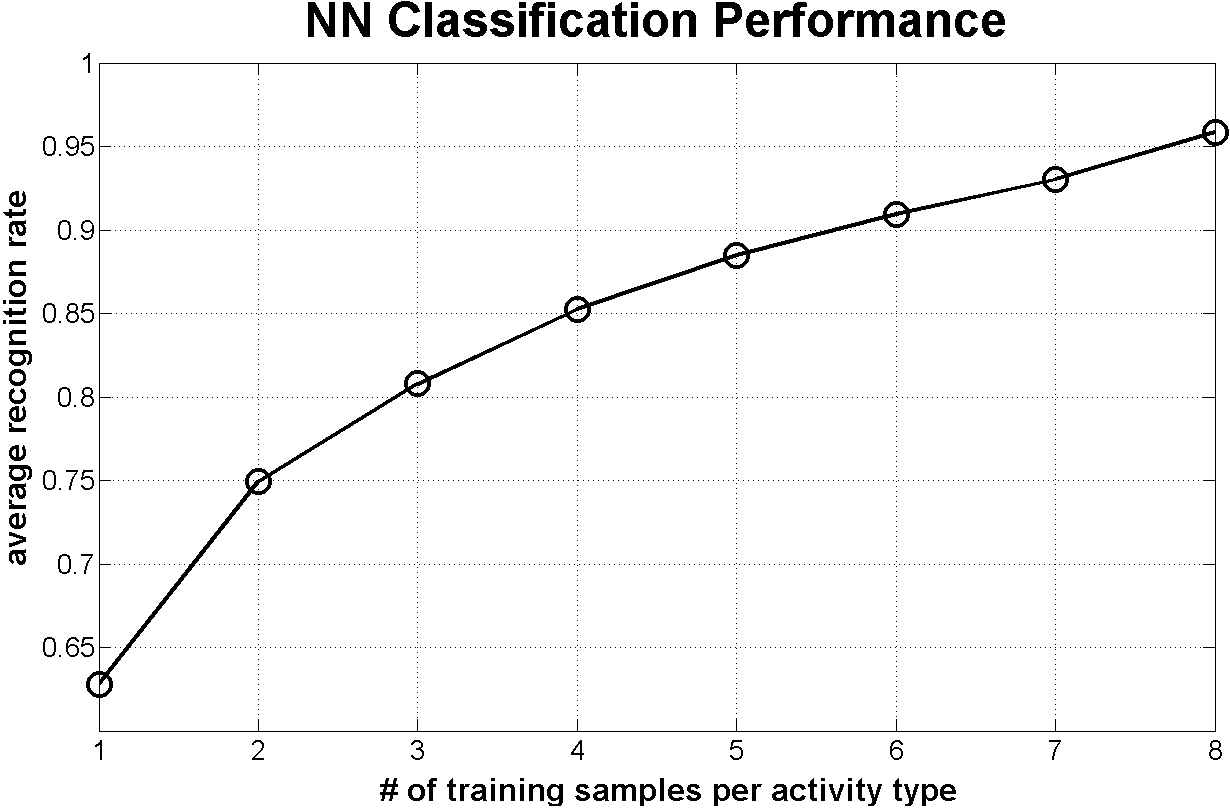} }  &   \includegraphics[width=.21\textwidth]{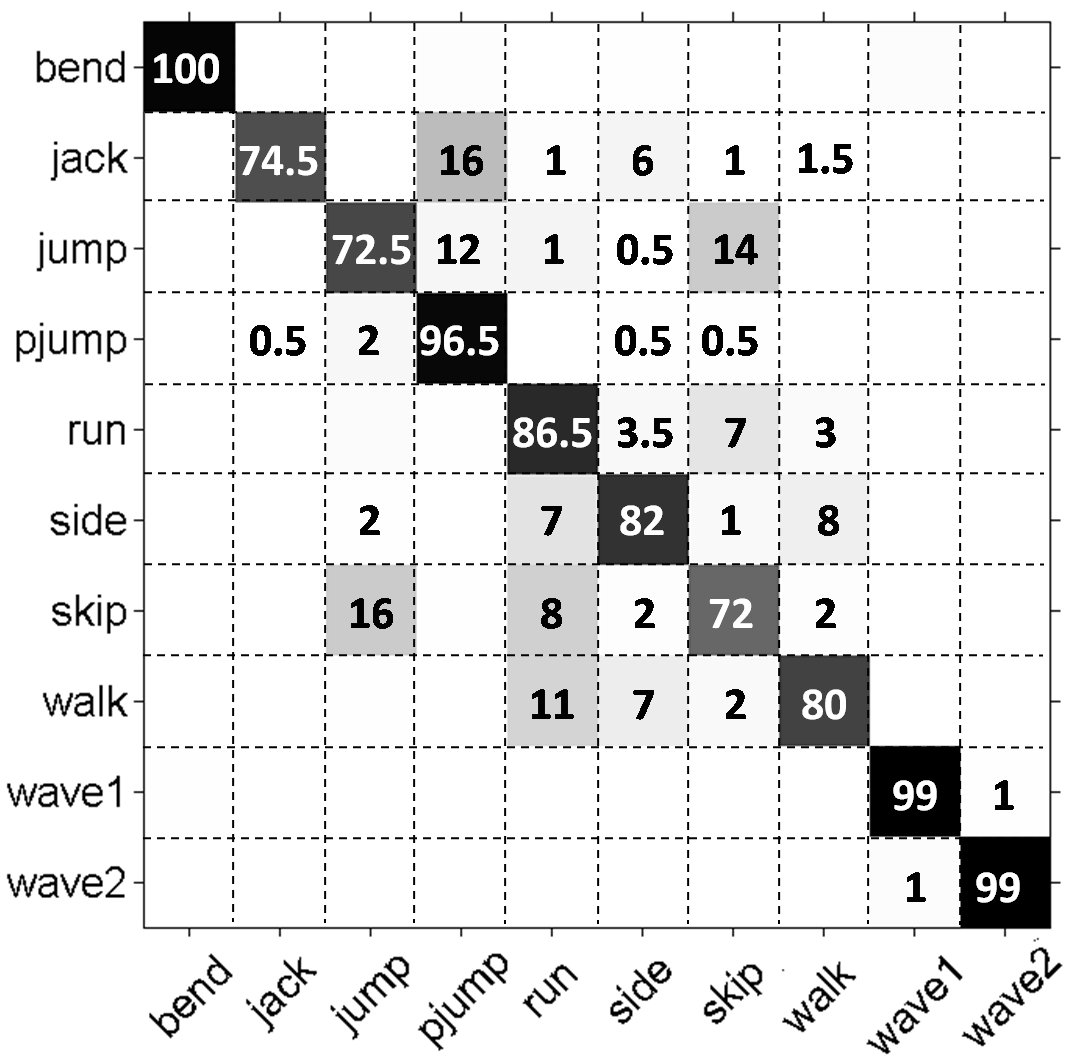}\\
          \textrm{(a) NN recognition performance } & \textrm{(b) confusion matrix}
     \end{array}$
     \end{narrow}

     \begin{narrow}{-1mm}{-3mm}
     \caption{\ref{fig: action recognition}(a) plots the recognition performance of a NN classifier vs. the number of training samples used per action type. \ref{fig: action recognition}(b) shows the confusion matrix. Darker squares indicate higher percentages.}\label{fig: action recognition} \end{narrow}
\end{figure}

After learning the texel transformations, we use a nearest neighbor (NN) classifier to recognize a test action sequence, given a set of training sequences. We define the dissimilarity between two sequences ($\mathcal{S}_1$ and $\mathcal{S}_2$) as the DTW (dynamic time warping) cost needed to warp the transformations of $\mathcal{S}_1$ into those of $\mathcal{S}_2$, where the dissimilarity between transformations $X_1$ and $X_2$ is defined as: $d\left(X_1,X_2\right)=1-\frac{\text{trace}\left(X_1^TX_2\right)}{\|X_1\|_F\|X_2\|_F}$. This cost is efficiently computed using dynamic programming. Figure \ref{fig: action recognition}(a) plots the variation of the average recognition rate versus the number of sequences (per action class) used for training. For each training sample size, we randomly choose a set of such size from each action class and perform classification. We repeat this multiple times and average the recognition rate to obtain the plotted values. Obviously, the performance improves as the number of training samples increases. More importantly, we note that a simple classifier using only one training sample achieves a $62\%$ recognition rate, where random chance is $10\%$.  Furthermore, Figure \ref{fig: action recognition}(b) shows the average confusion matrix. Note the high diagonal values. Here, we point out that confusion occurred between similar actions especially for the (``jump", ``skip") and (``run", ``walk") pairs. Better performance is expected, when texels are extracted more reliably and features are more discriminative of human motion.

\subsubsection{Multiple Swarm Element Sequences}
 We apply our algorithm to swarm video sequences compiled from online sources. We perform model learning and motion discrimination on four  sequences: ``birds" \cite{WangZhuIJCV05}, ``geese",  ``robot swarm" \cite{RobotSwarmDatabase}, and ``pedestrian" \cite{ChanPAMI2008}.

\vspace{2mm}
\noindent \textbf{Model Learning:} The features we used were based on a polar coordinate system centered at the centroid of each swarm element, where each angular bin had a width of $\frac{\pi}{10}$ rad. For each angular bin, we extracted two shape features (kurtosis and skew), the mean centroidal distance of the element boundary, and the mean intensity value. This yielded a feature vector of size $d=100$. Setting $\epsilon=10^{-3}$, $j_{\text{max}}=5$, $k_{\text{max}}=50$ and $W_T=5$, we applied \textbf{Algorithm \ref{algo: texel dynamics}} to learn the spatial layout and dynamics of each swarm sequence. To evaluate the performance of our method, we conducted a leave-five-out experiment, where we learn the swarm dynamics using all the frames except for five. The transformations and features of the elements in these left out frames are reconstructed using the AR model. We repeated this experiment and reported the average normalized residual errors in Table \ref{tab: residual error}, for the four sequences. These results show that our DS model represents the ground truth data well. Here, we note that the error was the highest for the ``pedestrian" sequence due to the variability in the swarm dynamics and appearance. Also, we compared these residual errors to the case when identity is used instead of the learned transformations (i.e. no dynamics update). The percentage ratio of these two errors are shown in parenthesis. We conclude that our learned dynamics substantially improve model fitting.

\begin{figure}[htp]
     \centering
     \begin{narrow}{-2mm}{0mm}
     $\begin{array}{cccc}
           \includegraphics[width=.105\textwidth]{ex_birds_rs.png}  &   \includegraphics[width=.105\textwidth]{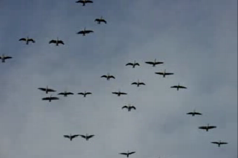} &
           \includegraphics[width=.105\textwidth]{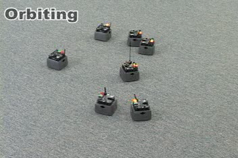}  &  \includegraphics[width=.105\textwidth]{ex_pedestrian.png}
          \end{array}$
          \end{narrow}
     \caption{``birds", ``geese",   ``robot", and ``pedestrian" swarms}\label{fig: experimental examples}
\end{figure}

\begin{table}[ht]
\begin{center}
\begin{tabular}[c]{|c|c|c|c|c|}
\hline
& ``birds" & ``geese" & ``robot" & ``pedestrian"\\
\hline
$e_R$ &  8.2 (5.4)   &  10.3 (4.9)    & 3.5 (4.2)     & 12.9 (9.5) \\
$e_S$ &  12.5 (6.8)  &  6.5 (5.8)       & 11.6 (5.5)  & 15.8 (11.6) \\
$e_T$ & 18.0 (4.1)  &  14.1 (7.7)   & 16.4 (4.4)   & 23.1 (18.3) \\ 
\hline
\end{tabular}
\end{center}
\caption{Average normalized residual error (as percentage). The percentage values in parentheses are the average errors normalized by the error incurred when the swarm dynamics are not updated.}\label{tab: residual error}
\end{table}

\vspace{2mm}
\noindent \textbf{Motion Discrimination:} Here, we demonstrate that our method can discriminate between different motions (i.e. sequences of transformations) within the same video sequence. After learning the swarm dynamics, we compute the dissimilarity in dynamics between every pair of swarm elements. We define the dissimilarity between two sequences of swarm element transformations ($\mathcal{T}_1$ and $\mathcal{T}_2$) as the dynamic time warping (DTW) cost needed to warp the transformations of $\mathcal{T}_1$ into those of $\mathcal{T}_2$ \cite{DTWReference}. Such a warping is crucial, since  $\mathcal{T}_1$ and  $\mathcal{T}_2$ might have different cardinalities (i.e. swarm elements \underline{do not} have to appear in the same number of frames). This DTW cost is efficiently computed using dynamic programming. However, to compute this sequence-to-sequence DTW cost, we need to define a distance between individual transformations comprising the sequences. We define the distance between transformations $X_1$ and $X_2$ as: $d\left(X_1,X_2\right)=1-\frac{\text{trace}\left(X_1^TX_2\right)}{\|X_1\|_F\|X_2\|_F}$. These DTW costs are employed in spectral clustering to cluster the swarm elements' dynamics.

\begin{figure}[htp]
     \centering
     \begin{narrow}{-3mm}{0mm}
     $\begin{array}{c}
          \includegraphics[width=.47\textwidth]{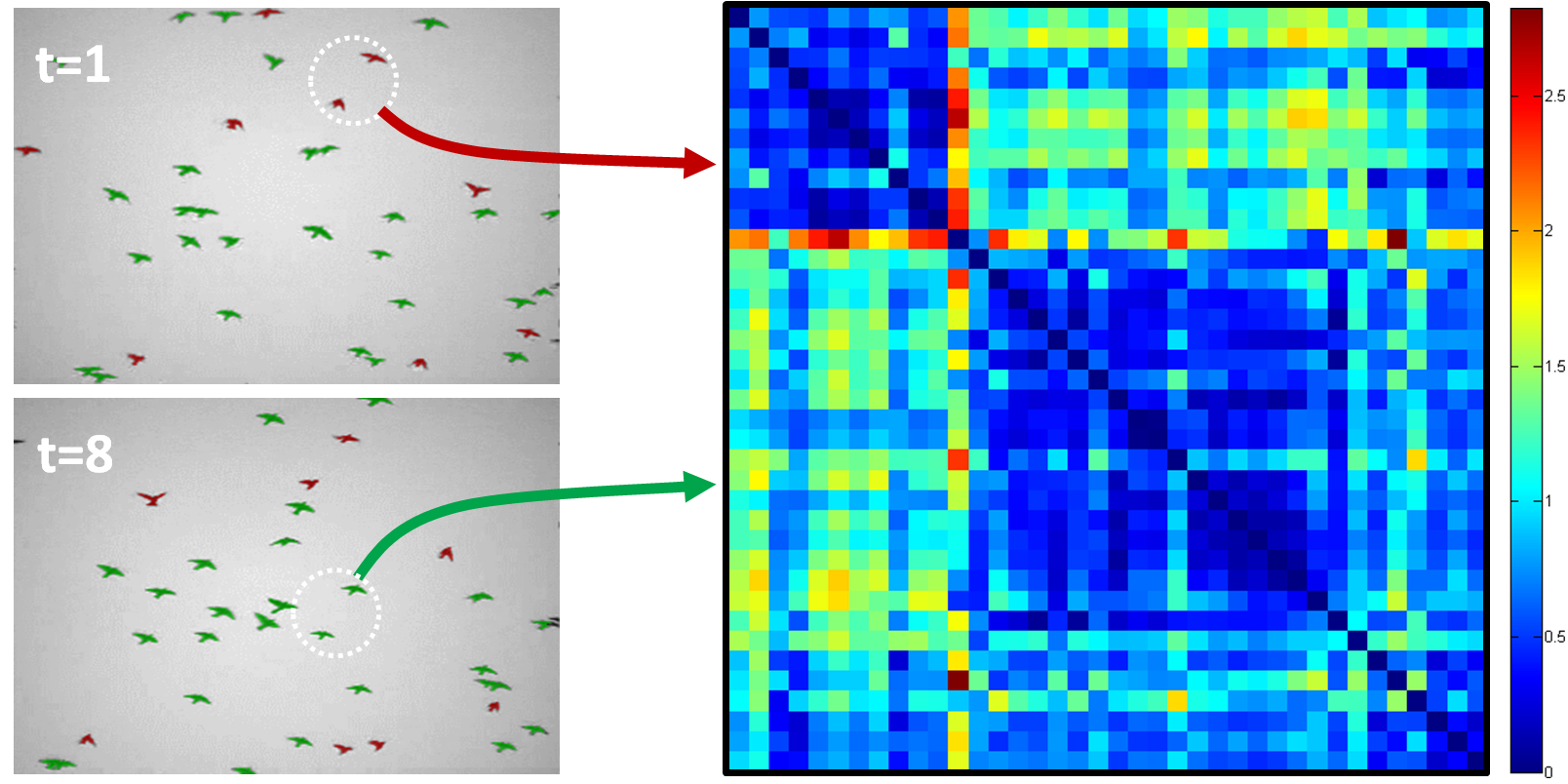}
          \end{array}$
          \end{narrow}
     \caption{Shows the ``birds" swarm example containing a ``bird-flapping" and ``bird-gliding" motion. The pairwise distances between the learned transformations are shown on the right. }\label{fig: birds discrimination}
\end{figure}

The ``birds" and ``pedestrian" sequences contain more than one distinguishable motion. \figlabel \ref{fig: birds discrimination} illustrates the clustering results obtained for the ``birds" sequence. The extracted swarm elements are color-coded in the frames according to their distinct motions. In this sequence, two types of motion co-exist: \textbf{(i)} a ``bird-flapping" motion where wings oscillate up and down and \textbf{(ii)} a ``bird-gliding" motion where the wings remain relatively still. On the right, \figlabel \ref{fig: birds discrimination} shows the DTW distances computed between all pairs of swarm element dynamics. We clearly see that type \textbf{(i)} elements undergo quite different transformations than those of type \textbf{(ii)}. Our approach was able to simultaneously learn the different dynamics in the sequence and discriminate them. This cannot be done by DT models such as \cite{WangZhuIJCV05}.

\begin{figure}
     \centering
          \includegraphics[width=.48\textwidth]{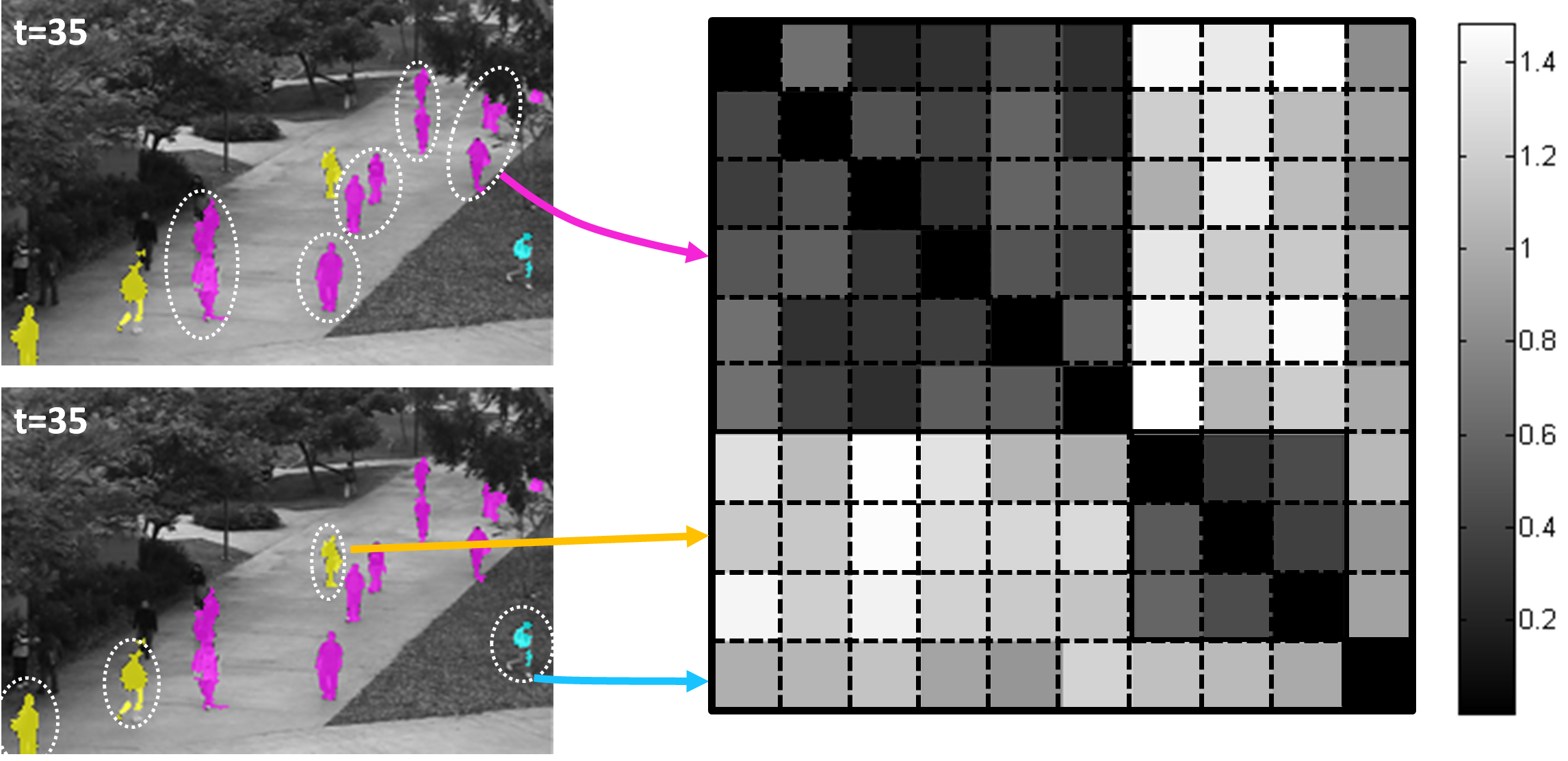}
     \caption{Shows a pedestrian example containing three types of motion. The extracted swarm elements are color-coded. The pairwise distances between the learned transformations are shown on the right. Brighter squares indicate larger distances. Refer to the \textbf{supplementary material} for these and other video results.}\label{fig: pedestrian}
\end{figure}

We also apply our algorithm to ``pedestrian" video sequences, where humans or groups of humans are considered swarm elements. These sequences were obtained from the UCSD pedestrian traffic database \cite{ChanPAMI2008}. Figure \ref{fig: pedestrian} illustrates the results obtained for a single pedestrian sequence that exhibits dense swarm activity. The extracted swarm elements are color-coded in the frames according to their distinctive dynamics. In this sequence, three types of motion co-exist. \textbf{(i)} Elements (some of which are groups of pedestrians) move/walk from the top right corner to the bottom left corner. \textbf{(ii)} Other elements moves in the opposite direction. \textbf{(iii)} One element represents a person crossing the grass instead of walking along the diagonal path. On the right, Figure \ref{fig: pedestrian} shows the DTW distances computed between all pairs of swarm elements. We see that the elements of \textbf{(i)} undergo much more similar transformations than those of \textbf{(ii)-(iii)}, which, in turn, have significantly different dynamics. Some pedestrian segments were not part of the spatial layout since they were indistinguishable from the background.

\section{Conclusion}
This paper proposes a spatiotemporal model for learning the spatial layout and dynamics of elements in swarm sequences. It represents a swarm element's motion as a sequence of linear transformations that reproduce its properties subject to local stationarity constraints. We conducted experiments on real sequences to demonstrate our approach's merit in representing swarm dynamics and discriminating between different dynamics. Our future goal is to apply this method to motion synthesis and recognition. The support of the Office of Naval Research under grant N00014-09-1-0017 and the National Science Foundation under grant IIS 08-12188 is gratefully acknowledged.

{
\bibliographystyle{ieee}
\bibliography{egbib_reduced}
}

\end{document}